\pdfoutput=1

\documentclass[11pt]{article}

\usepackage[]{ACL2023}

\usepackage{times}
\usepackage{latexsym}

\usepackage[T1]{fontenc}

\usepackage[utf8]{inputenc}

\usepackage{microtype}
\usepackage{times}
\usepackage{latexsym}
\usepackage{url}
\usepackage{hyperref}
\usepackage{graphicx}
\usepackage{amsmath}
\usepackage{amssymb}
\usepackage{subfigure}
\usepackage{caption}
\usepackage{algpseudocode}
\usepackage{multicol}
\usepackage{multirow}
\usepackage{booktabs}
\usepackage{listings}
\usepackage{array}
\usepackage{bm}
\usepackage{makecell}
\usepackage[ruled,linesnumbered]{algorithm2e}
\usepackage{inconsolata}

\title{Long-Tailed Question Answering in an Open World}

\author{
Yi Dai$^{1}$\thanks{\quad Work done while the author was interning at Alibaba.} \thanks{\quad Equal contribution.}, \
Hao Lang$^{2}$\footnotemark[2] \thanks{\quad \small Corresponding author.}, \
Yinhe Zheng$^{2}$, \
Fei Huang$^{2}$, \
Yongbin Li$^{2}$\footnotemark[3] \\
$^1$ Department of Computer Science and Technology, Tsinghua University
$^{2}$ Alibaba Group\\
\small \texttt{\{hao.lang, f.huang, shuide.lyb\}@alibaba-inc.com}, \\
\small \quad \texttt{dai-y21@mails.tsinghua.edu.cn, zhengyinhe1@163.com}\\
}

\begin{document}
\maketitle
\begin{abstract}

Real-world data often have an open long-tailed distribution, and building a unified QA model supporting various tasks is vital for practical QA applications.
However, it is non-trivial to extend previous QA approaches since they either require access to seen tasks of adequate samples or do not explicitly model samples from unseen tasks.
In this paper, we define Open Long-Tailed QA (OLTQA) as learning from long-tailed distributed data and optimizing performance over seen and unseen QA tasks.
We propose an OLTQA model that encourages knowledge sharing between head, tail and unseen tasks, and explicitly mines knowledge from a large pre-trained language model (LM).
Specifically, we organize our model through a pool of fine-grained components and dynamically combine these components for an input to facilitate knowledge sharing.
A retrieve-then-rerank frame is further introduced to select in-context examples, which guild the LM to generate text that express knowledge for QA tasks.
Moreover, a two-stage training approach is introduced to pre-train the framework by knowledge distillation (KD) from the LM and then jointly train the frame and a QA model through an adaptive mutual KD method.
On a large-scale OLTQA dataset we curate from 43 existing QA datasets, our model consistently outperforms the state-of-the-art. We release the code and data at \url{https://github.com/AlibabaResearch/DAMO-ConvAI/tree/main/oltqa}.

\end{abstract}

\section{Introduction}

Real-world data often have a long-tailed and open-ended distribution~\citep{liu2019large}.
As a cornerstone for AI applications~\cite{yang2019end}, Question Answering (QA) is widely investigated to tackle various QA tasks involving diverse formats and domains~\citep{khashabi-etal-2020-unifiedqa,zhong-etal-2022-proqa}.
The frequency distribution of QA tasks in our daily life is long-tailed~\citep{reed2001pareto}, with a few head tasks of adequate samples and many more tail tasks of limited samples, and we continuously encounter new tasks that are not seen during training in an open world.

We formally study \textit{Open Long-Tailed QA} (OLTQA) emerging in natural data settings.
A practical QA system shall learn from long-tailed distributed data, i.e., a few head tasks and many tail tasks, and it is expected to perform well over a balanced test set which include head, tail, and unseen tasks.

OLTQA must handle not only few-shot learning for tail tasks in the closed world~\citep{shu2017doc}, but also zero-shot learning for unseen tasks in an open world~\citep{scheirer2012toward} with one unified model.
A major challenge for OLTQA is the lack of knowledge required for the language understanding and reasoning abilities of QA tasks, especially under such low resource conditions~\citep{yan-etal-2020-multi-source}.
Therefore, it is important for an OLTQA model to share knowledge between head, tail, and unseen QA tasks~\citep{zaremoodi-etal-2018-adaptive}, and mine knowledge from external resources~\citep{,liu-etal-2022-generated}.\looseness=-1

However, it is non-trivial to directly extend previous methods to the OLTQA setting.
Specifically, an effective implementation of knowledge sharing is the multi-task learning (MTL) approach~\citep{liu-etal-2019-multi,raffel2020exploring}, in which task-specific components are maintained to preserve learned knowledge~\citep{aghajanyan-etal-2021-muppet,karimi-mahabadi-etal-2021-parameter}.
As we constantly encounter new tasks in practice, it is challenging to directly apply MTL methods since they do not explicitly model samples from unseen tasks.

Another challenge is the absence of samples from unseen tasks in the training process, which leads to poor prior knowledge about unseen tasks.
Fortunately, a large pre-trained language model (LM) embeds broad-coverage knowledge that can help a variety of tasks~\citep{rubin-etal-2022-learning}.
One key ingredient in LM knowledge mining is to select demonstrative in-context examples, which guild the LM to generate text that express knowledge for downstream tasks~\citep{liu-etal-2022-makes}.
However, few studies have explored selecting in-context examples to directly optimize QA performance in the OLTQA setting.

In this study, we propose an OLTQA model to address challenges mentioned above for the OLTQA setting.
Specifically, to encourage knowledge sharing between head and tail tasks while acknowledging the emergence of unseen tasks, we organize our model at the instance-level and use a dynamic architecture for each input~\citep{Wiwatcharakoses2020SOINNAS}, i.e., a pool of fine-grained components are maintained  and dynamically combined in each forward pass based on the input~\citep{Wang2021LearningTP}.
This scheme tackles unseen tasks, since the learned knowledge is distributed into different model components~\citep{Trauble2022DiscreteKB}.

We further mine knowledge from a large pre-trained LM.
Concretely, we employ a retrieve-then-rerank frame~\citep{ren-etal-2021-rocketqav2} to select demonstrative in-context examples for a test instance, which guide the LM to decode the output~\citep{brown2020language}.
The LM outputs are viewed as hints for QA tasks~\citep{zhang2022birdqa} and leveraged for improving QA performance.
The retrieve-then-rerank frame consists of an efficient retriever and an effective re-ranker~\citep{zamani2022stochastic}, which is optimized by a two-stage training approach.
The first stage pre-trains the retrieve-then-rerank framework by knowledge distillation from a pre-trained LM~\citep{Izacard2022FewshotLW}.
The second stage jointly train the above framework and an encoder-decoder QA model through adaptive mutual knowledge distillation~\citep{xie2022performance} to allow information exchange between each other.
Our key contributions are summarized as follows:

\begin{itemize}
    \item We formally define the OLTQA task, which learns from natural long-tail distributed data and optimizes the performance over seen and unseen tasks. We curate a large OLTQA dataset according to a
long-tail distribution from  43 existing representative QA datasets.
    \item We propose an OLTQA model, consisting of knowledge sharing and knowledge mining components to address challenges of OLTQA. An instance-level knowledge sharing mechanism is introduced, and a retrieve-then-rerank frame is employed to mine knowledge from a large pre-trained LM through a novel two-stage knowledge distillation training process.
    \item Our extensive experimentation on the OLTQA dataset demonstrates that our model consistently outperforms the state-of-the-art.
\end{itemize}

\section{Related Work}

\textbf{Question Answering (QA)} is important for advanced AI applications \cite{yang2019end}.
Recent approaches try to build unified QA models by casting different QA tasks into a unified text-to-text format~\cite{mccann2019the,khashabi-etal-2020-unifiedqa,zhong-etal-2022-proqa}.
Some works try to improve QA performance under the low-resource conditions~\citep{yan-etal-2020-multi-source,van2021cheap,bai2022domain}.
Some approaches also attempt to solve the open-domain QA problem, aiming at answering general domain questions through an extensive collection of documents~\cite{voorhees1999trec,chen-etal-2017-reading,NEURIPS2021_da3fde15,cheng-etal-2021-unitedqa}.
These approaches do not learn from natural long-tail distributed data.


\textbf{Long-Tailed Learning} focuses on long-tail distributed data~\citep{liu2019large}.
Recent approaches for long-tailed learning include re-balancing~\citep{Zhang_2021_ICCV}, information augmentation~\citep{He_2021_ICCV}, and module improvement~\citep{Cui_2021_ICCV}.
In this study, we attempt to build a QA model from long-tail distributed data by knowledge sharing and knowledge mining.


\textbf{Knowledge Mining} from external resources is essential for building robust QA models~\cite{pan-etal-2019-improving-question}.
Wikipedia and knowledge bases are used to improve QA performance~\citep{bi-etal-2019-incorporating,banerjee-etal-2019-careful}.
Large pre-trained LMs store rich knowledge, which is used to solve various tasks via conditioned generation~\cite{petroni-etal-2019-language}.
Recent approaches build prompt retrievers to select in-context examples from a training set to optimize LM generation performance~\citep{rubin-etal-2022-learning}.
However, these approaches cannot directly optimize our OLTQA model. In this study, we jointly train a retrieve-then-rerank framework and a QA model to enhance QA performance.

\textbf{Knowledge distillation (KD)} is often employed to learn a student model using the knowledge distilled from a teacher model by enforcing the agreement of outputs between the two models~\cite{hinton2015distilling}.
Mutual KD helps a group of models mutually generate knowledge to train each other~\citep{zhao2021novel}.
Our OLTQA model jointly trains the retrieve-then-rerank frame and the QA model through adaptive mutual KD, encouraging them to collaborate with each other~\citep{xie2022performance}.

\section{Method}

\subsection{Problem Setup}

In this study, we aim to learn from $n$ QA tasks $\{T_1,\cdots,T_n\}$, in which training sets follow a long-tailed Zipf distribution with power value $\alpha$, i.e., a few head tasks of adequate samples and many tail tasks of limited samples.
Each sample of $T_i$ is a tuple of a context $\boldsymbol{c}$, a question $\boldsymbol{q}$, and an answer $\boldsymbol{a}$: $\langle \boldsymbol{c}, \boldsymbol{q}, \boldsymbol{a} \rangle$.
Our QA model $F$ is built to predict $\boldsymbol{a}$ based on $\boldsymbol{c}$ and $\boldsymbol{q}$.
We also consider a more challenging setting in an open world, i.e., model $F$ needs to predict answers for unseen tasks.
Therefore, we collect another $\widetilde{n}$ unseen tasks $\{T_{n+1}, \cdots, T_{n+\widetilde{n}}\}$ that are only used for testing.

\begin{figure*}[t]
\centering
\includegraphics[width = 1.0\linewidth]{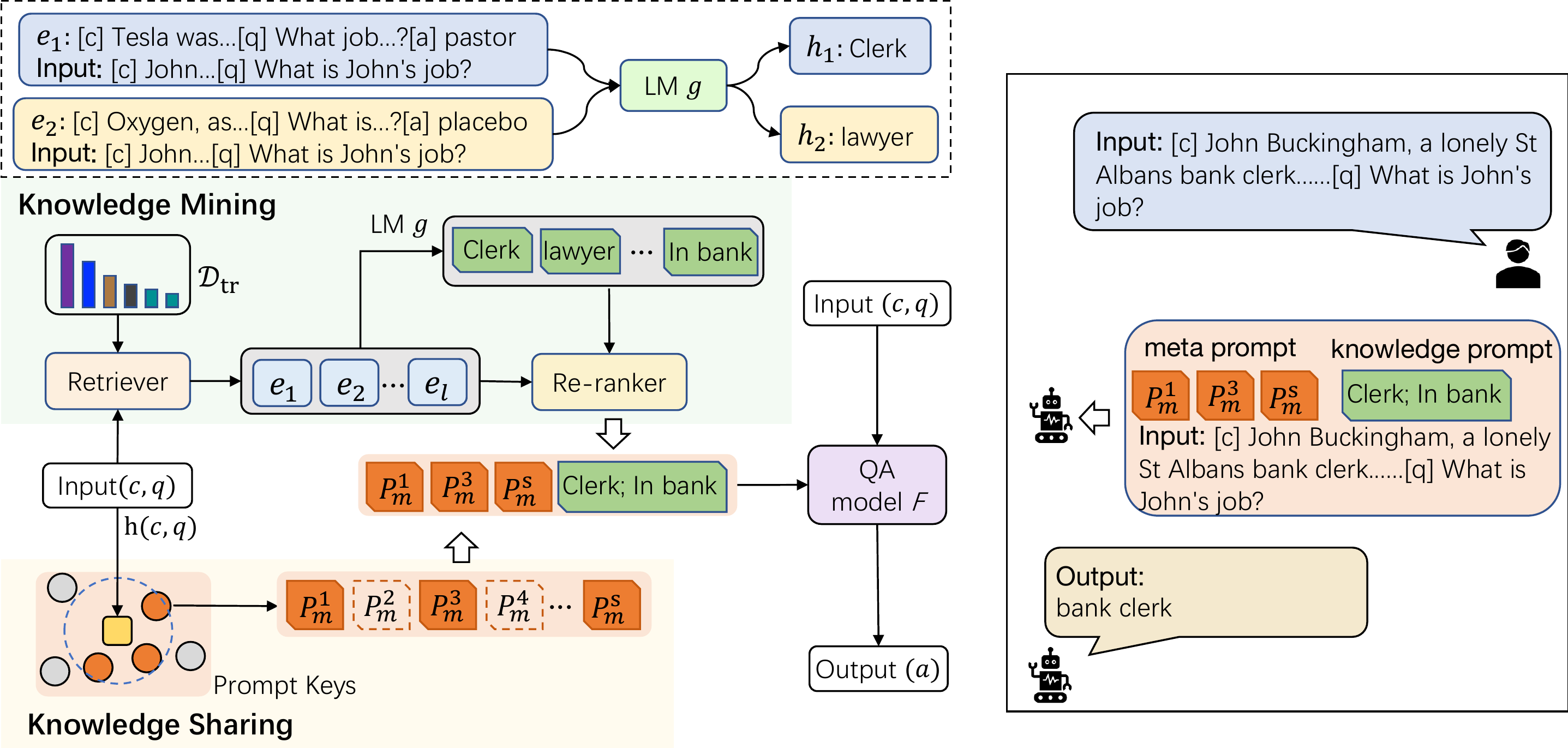}
\caption{Two key ingredients introduced in our model:(a) Knowledge sharing between head, tail, and unseen tasks at the instance level by maintaining a pool of prompts $\{\mathcal{P}_m^i\}_{i=1}^s$; (b) Knowledge mining from a pre-trained LM $g$ using a retrieve-then-rerank framework.}
\label{fig:pipeline}
\end{figure*} 

\subsection{Overview}
Our model tackles the open long-tailed QA problem by training a prompt-enhanced encoder-decoder QA model $F$ on long-tailed distributed data.
There are mainly two challenges to be addressed: (1) How to alleviate the low-resource problem and share knowledge between head, tail, and unseen tasks; (2) How to mine knowledge from external resources.
These two issues are tackled with two key ingredients in our model (see Figure~\ref{fig:pipeline}): 1. An instance-level knowledge sharing method (Section~\ref{sec:ks}); 2. A knowledge mining method from a pre-trained language model (Section~\ref{sec:km}).

We follow previous approaches to serialize the context $\boldsymbol{c}$, question $\boldsymbol{q}$, and answer $\boldsymbol{a}$ into text sequences~\cite{khashabi-etal-2020-unifiedqa,zhong2022proqa}.
For each training sample $\langle \boldsymbol{c}, \boldsymbol{q}, \boldsymbol{a} \rangle$, we first construct a prompt $\mathcal{P}$ based on $\boldsymbol{c}$ and $\boldsymbol{q}$, and then the encoder takes in the concatenation of $\mathcal{P}$, $\boldsymbol{c}$, and $\boldsymbol{q}$ and the decoder predicts $\boldsymbol{a}$, i.e., $p(\boldsymbol{a}|[\mathcal{P};\boldsymbol{c};\boldsymbol{q}])$, where $[;]$ denotes the sequence concatenation operation.\looseness=-1
~Specifically, $\mathcal{P}$ is a concatenation of two kinds of prompts, i.e., a meta prompt $\mathcal{P}_m$ and a knowledge prompt $\mathcal{P}_k$.
To capture fine-grained knowledge distributed in each input sample, we maintain $s$ meta prompts $\{\mathcal{P}_m^i\}_{i=1}^s$ and dynamically combine these prompts based on $\boldsymbol{c}$ and $\boldsymbol{q}$ to obtain $\mathcal{P}_m$~\cite{Wang2021LearningTP}.
We associate a key vector $\boldsymbol{k}_m^i$ for each meta prompt $\mathcal{P}_m^i$, respectively.
A fixed query function $h$ is built to map $\boldsymbol{c}$ and $\boldsymbol{q}$ to a query vector $\boldsymbol{x}=h(\boldsymbol{c}, \boldsymbol{q})$.
$h$ is initialized by a fixed pre-trained LM and not tuned in the training phase.
$\mathcal{P}_m$ can be determined by retrieving the most similar key vectors $\boldsymbol{k}_m^i$ using $\boldsymbol{x}$.
Note that $\mathcal{P}_m$ is a soft prompt, i.e., a sequence of trainable embeddings that is randomly initialized and optimized when training QA model $F$~\citep{liu2021gpt}.

We also mine knowledge from a large pre-trained LM $g$ to construct knowledge prompt $\mathcal{P}_k$.
\citet{liu-etal-2022-makes} showed that the efficacy of output generated by an LM could vary widely depending on the choice of in-context examples.
In this study, we introduce a retrieve-then-rerank framework $\langle R_1, R_2 \rangle$~\citep{ren-etal-2021-rocketqav2} to select in-context examples from a training set $\mathcal{D}_{tr}$, consisting of a retriever $R_1$ and a re-ranker $R_2$~\citep{zamani2022stochastic}.
The retriever $R_1$ is implemented as an efficient dual-encoder~\cite{xiong2021approximate}.
The re-ranker $R_2$ is built as a more effective cross-encoder~\cite{luan-etal-2021-sparse}.
For a test instance $\langle \boldsymbol{c}, \boldsymbol{q} \rangle$, we mine knowledge following three steps:
\textbf{1.} $R_1$ retrieves a subset of $l$ candidate examples $\{\boldsymbol{e}_i=\langle \boldsymbol{c}_i, \boldsymbol{q}_i, \boldsymbol{a}_i \rangle\}_{i=1}^l$ from training set $\mathcal{D}_{tr}$;
\textbf{2.} LM $g$ produces a text $\boldsymbol{h}_i$ for each example $\boldsymbol{e}_i$ by conditional generation $p_g(\boldsymbol{h}_i|[\boldsymbol{e}_i;\boldsymbol{c};\boldsymbol{q}])$, which can serve as a hint for the test instance;
\textbf{3.} $R_2$ further select top $\widetilde{l}$ hints $\{\boldsymbol{h}_i\}_{i=1}^{\widetilde{l}}$ to obtain the knowledge prompt $\mathcal{P}_k$ $(\widetilde{l} \ll l)$, in which the scoring function measures the similarity between $\langle \boldsymbol{c}, \boldsymbol{q} \rangle$ and $\langle\boldsymbol{e}_i,\boldsymbol{h}_i \rangle$.
Note that $\mathcal{P}_k$ is a hard prompt~\cite{jiang-etal-2020-know}, which is a concatenation of texts in $\{\boldsymbol{h}_i\}_{i=1}^{\widetilde{l}}$.

\subsection{Instance-level Knowledge Sharing}\label{sec:ks}
To facilitate knowledge sharing between head, tail, and unseen tasks at the instance level, we maintain a pool of prompts and optimize key vectors assigned to these prompts.
Specifically, for each input $\langle \boldsymbol{c},\boldsymbol{q} \rangle$, we select $\widetilde{s}$ prompt keys that are closest to the query vector $\boldsymbol{x}=h(\boldsymbol{c},\boldsymbol{q})$ and concatenate these $\widetilde{s}$ associated meta prompts to obtain $\mathcal{P}_m$.
Intuitively, the knowledge associated with the input sample is distributed in these $\widetilde{s}$ meta prompts.

When learning meta prompt keys, we assume the distribution of these keys should balance  diversity and locality.
Concretely, meta prompts are expected to distribute to the whole vector space so that every meta prompt can be involved in the training process, while similar prompt keys are grouped into clusters so that the knowledge of each sample can be better shared.
We propose  the following loss to enforce the above two properties:
\begin{equation}
\begin{aligned}
\label{eq:meta}
    \mathcal{L}_{m} = &  \mathop{\mathbb{E}}_{\langle \boldsymbol{c}, \boldsymbol{q}, \boldsymbol{a} \rangle \in \mathcal{D}_{tr} }(  \mathop{\sum}\limits_{i \in \mathcal{S}(\boldsymbol{x})}{\rm  max}(0,||\bm{k}_m^{i}, \boldsymbol{x}||-\eta)\\&+\mathop{\sum}\limits_{i, j \in \mathcal{S}(\bm{x})}{\rm max}(0,\gamma-||\bm{k}_m^{i},\bm{k}_m^{j}||)/{\widetilde{s}}^2),
\end{aligned}
\end{equation}
where the operator $||\cdot,\cdot||$ determines the distance between two input vectors (here we use cosine distance), $\mathcal{D}_{tr}$ is the training set of all seen tasks, $\mathcal{S}(\boldsymbol{x})$  is the index set of $\widetilde{s}$ selected meta prompt keys that are closest to $\boldsymbol{x}$,
$\eta$ and $\gamma$ are scalar hyper-parameters to control the distance margin. Specifically, the first term in the above equation pulls these selected meta prompt keys around the query vector. The second term pushes these keys away from each other to occupy the whole vector space.

\subsection{Pre-trained LM Knowledge Mining}\label{sec:km}

To further enhance QA performance, we also mine knowledge from a large pre-trained LM $g$.
We employ a retrieve-then-rerank framework $\langle R_1, R_2 \rangle$ to retrieve in-context examples from a training set $\mathcal{D}_{tr}$ and further select hints for the test instance that are generated by LM $g$.
We propose a two-stage knowledge distillation method to jointly train the framework $\langle R_1, R_2 \rangle$ and QA model $F$.

\paragraph{Stage \uppercase\expandafter{\romannumeral1}.} We pre-train $R_1$ and $R_2$ by knowledge distillation from a pre-trained LM $g$, inspired by ~\citet{rubin-etal-2022-learning}.
We first construct a set of $c$ candidate examples $\{\boldsymbol{e}_i=\langle \boldsymbol{c}_i, \boldsymbol{q}_i, \boldsymbol{a}_i \rangle\}_{i=1}^c$ for a traning instance $\langle \boldsymbol{c}, \boldsymbol{q},\boldsymbol{a} \rangle$ with BM25~\citep{robertson2009probabilistic} .
Then, we score each candidate example $\boldsymbol{e}_i$ and calculate a distribution of candidate examples by applying the Softmax operator over the resulting scores, based on scoring functions of LM $g$, $R_1$, and $R_2$, respectively.
Specifically, the distribution for the LM  $g$ scoring function is:
\begin{equation}
p_{lm}(\boldsymbol{e}_k)=\frac{\exp(\log(p_{g}(\boldsymbol{a}|[\boldsymbol{e}_k; \boldsymbol{c}; \boldsymbol{q} ])))}{\sum_{i=1}^c \exp(\log(p_{g}(\boldsymbol{a}|[\boldsymbol{e}_i; \boldsymbol{c}; \boldsymbol{q} ])))}, \nonumber
\end{equation}
where $p_{g}(\boldsymbol{a}|[\boldsymbol{e}_k;\boldsymbol{c}; \boldsymbol{q}])$ is the score for candidate $\boldsymbol{e}_k$, which is the probability under LM $g$ of output sequence conditioned on the candidate example and the training instance.
In a similar manner, we calculate distributions $p_{r1}$ and  $p_{r2}$ based on scoring functions of $R_{1}$ and $R_{2}$, respectively.
We optimize $R_1$ and $R_2$ by minimizing KL-divergence of $p_{lm}$ from $p_{r1}$ and $p_{r2}$~\citep{Izacard2022FewshotLW}:

\begin{equation}
\label{s1}
    \begin{aligned}
    \mathcal{L}_{lm} = \mathop{\mathbb{E}}\limits_{\langle \boldsymbol{c}, \boldsymbol{q},\boldsymbol{a} \rangle \in \mathcal{D}_{tr}}&({\rm KL}(\dashv[p_{lm}]\|p_{r1})\\&+{\rm KL}(\dashv[p_{lm}]\|p_{r2})),
    \end{aligned}
\end{equation}
 where $\dashv[\cdot]$ is a stopgrad operator that sets the gradient of its operand to zero.

\paragraph{Stage \uppercase\expandafter{\romannumeral2}.}
We jointly train $\langle R_1$, $R_2 \rangle$ and the QA model $F$.
For each training sample $\langle \bm{c},\bm{q},\bm{a} \rangle$, we first construct prompt $\mathcal{P}_m$ and $\mathcal{P}_k$, and then optimize the encoder-decoder QA model $F$ together with $\mathcal{P}_m$ using the following loss:

\begin{equation}
\label{eq:qa}
\mathcal{L}_{f} = \mathop{\mathbb{E}}_{\langle \boldsymbol{c}, \boldsymbol{q},\boldsymbol{a} \rangle \in \mathcal{D}_{tr} }(-\log~p_{F}(\boldsymbol{a}|[\mathcal{P}_m;\mathcal{P}_k;\boldsymbol{c};\boldsymbol{q}])).
\end{equation}

To allow information exchange and encourage agreement between $\langle R_1, R_2 \rangle$ and QA model $F$, mutual knowledge distillation is introduced to refine $R_1$, $R_2$, and $F$ by knowledge distillation from each other~\citep{zhao2021novel}.
However, in this case, a worse-performing model is allowed to generate knowledge to train a better-performing model, which may lead to collective failures~\cite{xie2022performance}.
Therefore, we propose an adaptive mutual knowledge distillation method to allow a model to generate knowledge for training another model only if it performs better.

Therefore, we evaluate the performance of $R_1$, $R_2$, and $F$ on a validation set $\mathcal{D}_{val}$ before mutual knowledge distillation.
Specifically, we select top $\widetilde{l}$ hints $\{\boldsymbol{h}_i\}_{i=1}^{\widetilde{l}}$ from the $c$ candidate examples $\{\bm{e}_i\}_{i=1}^c$ of a validation instance $\langle \boldsymbol{c}, \boldsymbol{q},\boldsymbol{a} \rangle$ based on scoring functions of $R_1$, $R_2$, $F$, and then obtain knowledge prompt $\mathcal{P}_k^{r1}$, $\mathcal{P}_k^{r2}$ and $\mathcal{P}_k^{f}$, respectively.
The scoring function of QA model $F$ is $p_{F}(\bm{a}|[\mathcal{P}_m;\bm{h}_i;\bm{c}; \bm{q}])$, where $\bm{h}_i$ is a hint for example $\bm{e}_i$ and acts as a pseudo knowledge prompt.
We evaluate $R_1$, $R_2$, and $F$ as follows:
\begin{equation}
\label{evl}
v_{i} =  \mathop{\mathbb{E}}_{\langle \boldsymbol{c}, \boldsymbol{q},\boldsymbol{a} \rangle \in \mathcal{D}_{val} }\log p_{F}(\boldsymbol{a}|[\mathcal{P}_m;\mathcal{P}_k^i; \boldsymbol{c};\boldsymbol{q} ]),
\end{equation}
where $i \in \{r1,r2,f\}$ denotes a specific model.
Lastly, we optimize the adaptive mutual knowledge distillation loss as follows:
\begin{equation}\small
\begin{aligned}
\label{mkd}
    \mathcal{L}_{mkd} = \mathop{\mathbb{E}}_{\langle \boldsymbol{c}, \boldsymbol{q},\boldsymbol{a} \rangle \in \mathcal{D}_{tr} }&\mathop{\sum}\limits_{i,j \in \{r1,r2,f\}}
    {\rm KL}(\dashv[p_i]\|p_j)\cdot\mathbb{I}(v_i>v_j),
\end{aligned}
\end{equation}
where $p_{f}$ is the distribution of candidate examples based on the scoring function of QA model $F$.

The whole training process of our model is summarized in Algorithm \ref{alg:algorithm}.

\begin{algorithm}
\caption{The training process}
\label{alg:algorithm}
\KwIn{Training data $\mathcal{D}_{tr}$, validation data $\mathcal{D}_{val}$.}
\KwOut{QA model $F$, meta prompts $\{\mathcal{P}_m^i\}_{i=1}^s$, prompt keys $\{\bm{k}_m^i\}_{i=1}^s$, framework $\langle R_1, R_2 \rangle$.}
\tcp{Stage \uppercase\expandafter{\romannumeral1}}
Train $R_1$ and $R_2$ using $\mathcal{L}_{lm}$ (Eq. \ref{s1}). \\
\tcp{Stage \uppercase\expandafter{\romannumeral2}}
Train $\{\bm{k}_m^i\}_{i=1}^s$ using $\mathcal{L}_{m}$ (Eq. \ref{eq:meta}). \\
Train $F$ and $\{\mathcal{P}_m^i\}_{i=1}^s$  using $\mathcal{L}_{f}$ (Eq. \ref{eq:qa}). \\
Evaluate $R_1$, $R_2$ and $F$ (Eq. \ref{evl}). \\
Train $R_1$, $R_2$, $F$, $\{\mathcal{P}_m^i\}_{i=1}^s$  using $\mathcal{L}_{mkd}$ (Eq. \ref{mkd}).\\
\end{algorithm}

\section{Experiments}

\subsection{Datasets}

\begin{figure}[!t]
\scalebox{0.85}{
\centering
\includegraphics[width=1.0\linewidth]{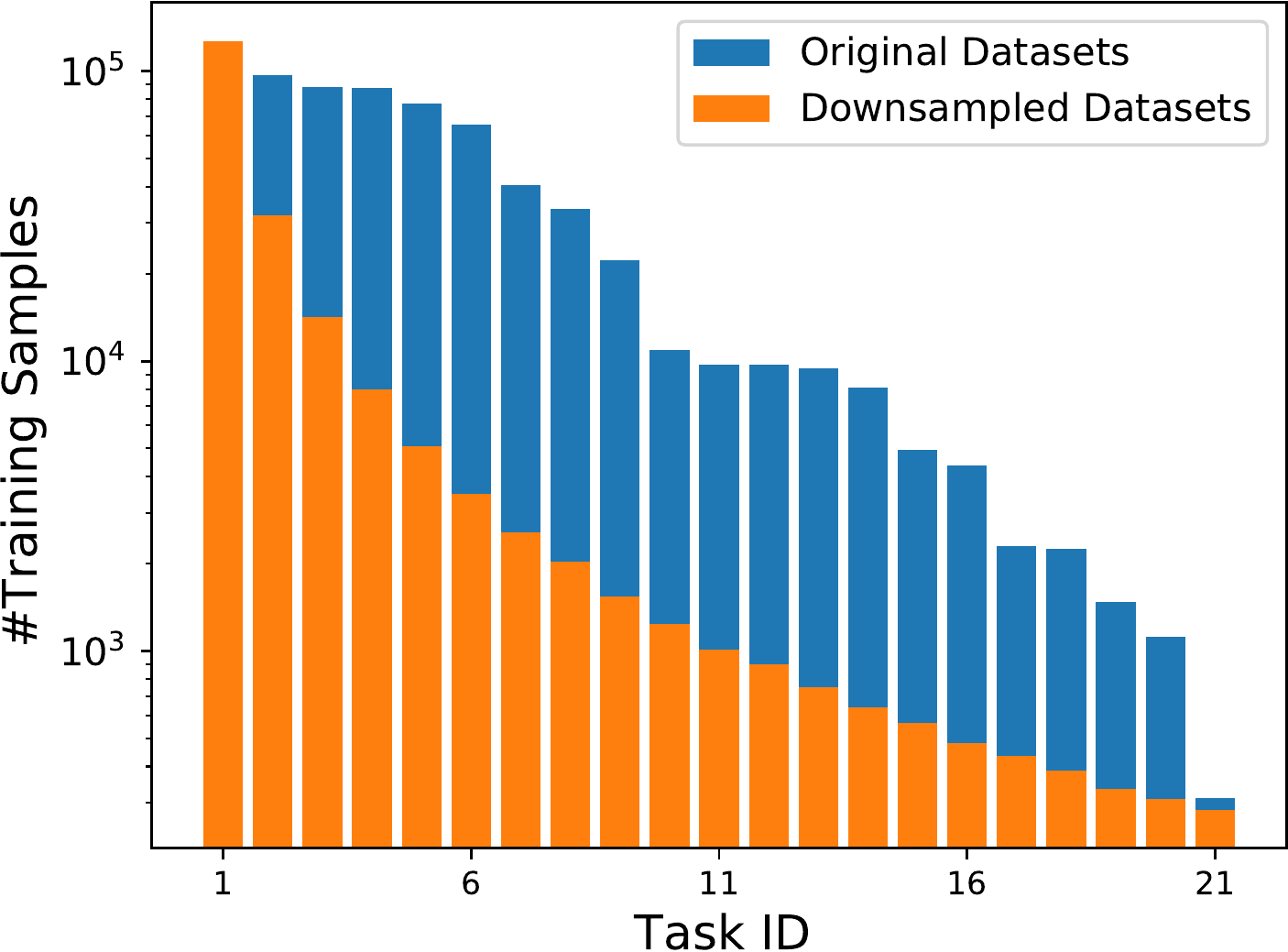}
}
\centering
\caption{Training dataset statistics of long-tailed QA tasks. Blue bars represent the original dataset sizes of 21 seen tasks and orange bars denote down-sampled dataset sizes.}

\label{fig:task-distribution}
\end{figure}

We curate an open long-tailed question answering benchmark from 43 existing representative QA datasets~\cite{unifiedqav2} covering four QA formats (\textit{Extractive} QA, \textit{Abstractive} QA, \textit{Multiple-choice} QA, and \textit{Yes/No} QA). See Appendix~\ref{append:datasets} for more details of the datasets.
We regard each dataset as an individual QA task and reserve $\widetilde{n}=22$ as unseen tasks. 
Our model is trained on the rest of $n=21$ seen tasks while tested on all 43 tasks.
We down-sample the training sets of all seen tasks following a Zipf distribution with power value $\alpha=2.0$ to construct the training data for our model.
Figure~\ref{fig:task-distribution} shows the training data statistics.

\label{sec:seen}
\begin{table*}[!t]
\centering
\small
\scalebox{0.9}{
\begin{tabular}{l|ccccccc|ccc}
\toprule
Methods & SQuAD 2 & NatQA & RACE & ARC-easy & MCTest & ARC-hard & MultiRC & Head@3 & Tail@4 & $A_{\rm{seen}}$\\
\midrule
UnifiedQA & 77.80 & 40.25 & 56.97 & 36.84 & 77.19 & 31.77 & 80.45 & 58.34 & 56.56 & 55.21\\
ProQA & 79.84 & 39.01 & 59.55 & 44.21 & 80.00 & 38.13 & 77.56 & 59.47 & 59.98 & 53.23 \\
Muppet & 79.41 & 40.83 & 57.13 & 38.07  & 79.06 & 31.34 & 85.57 & 59.12 & 58.51 & 56.13 \\
Hyperformer++ & 79.52 & 40.24 & 58.24 & 40.18 & 76.88 & 31.10 & 86.86 & 59.33 & 58.76 & 56.81  \\
EPR  & 44.14 & 39.50 & 38.82 & 51.81 & 55.00 & 39.80 & 56.41 & 40.82 & 50.76 & 47.97 \\
\midrule
Ours (w/o $\mathcal{P}_m$) & 77.72 & 42.10 & 58.13 & 56.49 & 83.02
& 39.46 & 85.58 & 59.32 & 66.14 & 59.60 \\
Ours (w/o $\mathcal{P}_k$) & 78.89 & 40.20 & 59.34 & 39.82 & 76.25
& 33.11 & 85.90 & 59.48 & 58.77 & 56.51 \\
Ours & \textbf{79.99} & \textbf{42.68} & \textbf{59.65} & \textbf{58.95} & \textbf{83.75} & \textbf{40.43} & \textbf{87.82} & \textbf{60.77} & \textbf{67.74} & \textbf{61.48} 
\\
\bottomrule
\end{tabular}}
\caption{Comparison with competitive baselines and ablations on main components of our model in seven seen tasks (3 head tasks + 4 tail tasks). Bold numbers are superior results.}
\label{tab:in-domain}
\end{table*}

\begin{table*}[!t]
\centering
\small
\scalebox{0.9}{
\begin{tabular}{l|ccccccc|c}
\toprule
\multirow{2}{*}{Methods}  & AdversarialQA & \multirow{2}{*}{RACE-C} & \multirow{2}{*}{MMMLU} & OneStopQA & \multirow{2}{*}{MCScript} & \multirow{2}{*}{DREAM} & \multirow{2}{*}{PubmedQA} & \multirow{2}{*}{$A_{\rm{unseen}}$}  \\
& dRoberta & &  & Advanced & & &  &\\
\midrule
UnifiedQA  & 18.16 & 49.86  & 28.77 & 54.01 & 67.97 & 59.56 & 50.53 & 46.70 \\
ProQA  & 14.21  & 54.91  & 25.96 & 61.11 & 71.23 & 64.41  & 58.00 & 48.27 \\
Muppet  & 17.33 & 50.00 & 30.42 & 54.79 & 70.91 & 58.61 & 56.73 & 46.98 \\
Hyperformer++  & 16.99  & 52.11  & 25.26 & 59.88 & 71.51 & 59.31 & 53.00 & 47.21 \\
EPR  & 27.74 & 35.39 & 28.77 & 60.49 & 65.56 & 53.92 & 59.67 & 46.57 \\
\midrule
Ours (w/o $\mathcal{P}_m$) & 25.16 & 53.51 & 33.68 & 61.11 & 77.46 & 68.28 & 62.07 & 52.09 \\
Ours (w/o $\mathcal{P}_k$)  & 17.12  & 53.23 & 31.23 & 56.70 & 70.80 & 60.29 & 56.27 & 48.37 \\
Ours  & \textbf{28.05}  & \textbf{56.88}  & \textbf{36.14} & \textbf{64.31} & \textbf{79.16} & \textbf{69.51} & \textbf{64.40} & \textbf{54.42} \\
\bottomrule
\end{tabular}}
\caption{Comparison with competitive baselines and ablations on main components of our model in seven unseen tasks (randomly selected). Bold numbers are superior results.}
\label{tab:out-domain}
\end{table*}

\subsection{Metrics}
The evaluation metric of each above task follows~\citet{unifiedqav2} (see more details in Appendix~\ref{append:datasets}).
We calculate the average performances over 21 seen tasks ($A_{\rm{seen}}$) and 22 unseen tasks ($A_{\rm{unseen}}$) to evaluate the QA performance.
We also calculate the average scores over a subset of seen tasks with $m$ largest training sets (Head@m) and $n$ smallest training sets (Tail@n) to evaluate the performance of head and tail tasks, respectively.

\subsection{Implementation Details}
 
We use T5-base~\cite{raffel2020exploring} to initialize the QA model $F$. For knowledge sharing, we maintain totally $s=30$ meta prompts, and set the length of each meta prompt to $10$. We adopt a fixed T5-base encoder with an average pooling layer to generate the query vector. For each instance, we select $\widetilde{s}=5$ meta prompts to construct $\mathcal{P}_m$. We set $\eta = 0.15$ and $\gamma = 0.3$ in Eq.~\ref{eq:meta}. For knowledge mining, we use a dual-encoder as retriever, and a cross-encoder as re-ranker. Encoders in the retriever and the re-ranker are all initialized with Bert-base-uncased~\cite{devlin-etal-2019-bert}. We use GLM-10B~\cite{du-etal-2022-glm} with 10B parameters as pre-trained LM $g$. For 
 each instance, the retriever first selects $l=64$ examples from the training dataset, and the re-ranker selects $\widetilde{l}=4$ examples to construct $\mathcal{P}_k$. All hyper-parameters are tuned according to the average score on the validation set. All results reported in our paper are averages of 3 runs with different random seeds. We use the AdamW~\cite{loshchilov2017decoupled} optimizer with a learning rate of 1e-4 and batch size of 32. Our model is trained for five epochs. All experiments are performed on 8 A100 GPUs.
See Appendix~\ref{app:details} for more implementation details. 

\subsection{Baselines}

We use the following competitive baselines:
\textbf{1. UnifiedQA}:~\cite{khashabi-etal-2020-unifiedqa} casts different QA tasks into a unified text-to-text format and builds a single model for all QA tasks;
\textbf{2. ProQA}:~\cite{zhong-etal-2022-proqa} uses structural prompts to train a unified QA model with a QA-centric pre-training;
\textbf{3. Muppet}:~\cite{aghajanyan-etal-2021-muppet} maintains task-specific heads and learns QA tasks through multi-task learning;
\textbf{4. Hyperformer++}:~\cite{karimi-mahabadi-etal-2021-parameter} uses a hyper-network to generate task-specific adapters for multi-task learning;
\textbf{5. EPR}:~\cite{rubin-etal-2022-learning} propose an efficient method to retrieve in-context examples for a test instance and use a pre-trained LM to directly decode the output based on the examples.
Note that ``Muppet'' and ``Hyperformer++'' have no specific modules for unseen tasks. Thus, we select a task with the lowest perplexity across all seen tasks for an input from unseen tasks in the testing phase, following~\citet{madotto-etal-2021-continual}.

\subsection{Main Results}

Table~\ref{tab:in-domain} shows the result on seen tasks.
Our model outperforms all competitive baselines in terms of Head@3, Tail@4, $A_{\rm{seen}}$, and achieves SOTA results on all head and tail tasks.
We can observe that:
\textbf{1.} Our model achieves an even larger performance improvement for tail tasks, i.e., absolute improvement is $1.44$ in Head@3 and $8.98$ in Tail@4, compared to the best-performing baseline Hyperformer++.
The performance gain precisely demonstrates the advantages of knowledge sharing between head and tail tasks and knowledge mining from external resources.
\textbf{2.} Our model also outperforms the in-context learning baseline EPR without any parameter update of the pre-trained LM.
This shows that leveraging knowledge mined from a pre-trained LM and directly optimizing QA tasks can lead to better QA performance. 
See Appendix~\ref{ap:or} for more evaluation details of all 21 seen tasks. 

Table~\ref{tab:out-domain} shows the result on unseen tasks.
Our model yields the best performances on all metrics.
We can also observe that:
\textbf{1.} Our model that shares knowledge through fine-grained components (i.e., a pool of meta prompts) and mines knowledge from an LM generally obtain higher performance.
\textbf{2.} EPR is on par with the other baselines trained on seen tasks. It shows that a pre-trained LM embeds a large amount of knowledge, which can help QA tasks potentially.

\subsection{Ablation Studies}

\paragraph{Model Main Components:}
Ablation studies are carried out to validate the effectiveness of each main component in our model.
Specifically, the following variants are investigated:
\textbf{1. w/o $\mathcal{P}_m$} removes the knowledge sharing component, i.e., meta prompt $\mathcal{P}_m$ is not used.
\textbf{2. w/o $\mathcal{P}_k$} removes the knowledge mining component, i.e., knowledge prompt $\mathcal{P}_k$ is not used.
Results in Table~\ref{tab:in-domain} and Table~\ref{tab:out-domain} indicate that our model outperforms all ablation variants.
Specifically, we can also observe that:
1. Both knowledge sharing (see w/o $\mathcal{P}_m$) and knowledge mining (see w/o $\mathcal{P}_k$) components help to improve the QA performance.
2. Knowledge mining brings larger improvement compared to knowledge sharing component on both tail and unseen tasks.  This further proves the importance of leveraging knowledge embedded in the pre-trained LM for the OLTQA setting. 
We provide examples where our model is correct and the variant without knowledge mining (i.e., w/o $\mathcal{P}_k$) is incorrect, together with 4 top hints selected by the retrieve-then-rerank framework in Appendix~\ref{apc:case}.

\begin{table}[!t]
\centering
\small
\begin{tabular}{c|l|c|c}
\toprule
     Categories & Variants & $A_{\rm{seen}}$ & $A_{\rm{unseen}}$\\
\midrule
\multirow{2}{*}{Retriever} 
& BM25 Retriever & 58.06 & 51.44  \\
& EPR Retriever  & 59.24 & 52.14  \\
\midrule
Re-ranker & w/o Re-ranker & 58.41 & 51.01\\
\midrule
\multirow{3}{*}{\shortstack{Knowledge\\Distillation}}
& w/o MKD  & 59.82  & 50.90 \\
& Static MKD   & 60.09 & 51.88 \\
& Back KD  & 60.21 & 52.35 \\
\midrule
\multicolumn{2}{c|}{Ours} & \textbf{61.48} & \textbf{54.42} \\
\bottomrule
\end{tabular}
\caption{Ablation on knowledge mining components.}
\label{tab:ablation}
\end{table}

\paragraph{Knowledge Mining Components:}
To evaluate design choices of retrieve-then-rerank framework $\langle R_1, R_2 \rangle$ and two-stage knowledge distillation (KD) in knowledge mining, we perform ablation on alternatives:
\textbf{1. BM25 Retriever} uses the unsupervised retriever BM25~\cite{robertson2009probabilistic} to replace retriever $R_1$.
\textbf{2. EPR Retriever} trains $R_1$ by using a pre-trained LM as the scoring function~\citep{rubin-etal-2022-learning}.
\textbf{3. w/o Re-ranker} removes the re-ranker $R_2$, and directly uses $R_1$ to select examples and generate hints.
\textbf{4. w/o MKD} removes the adaptive mutual KD loss $\mathcal{L}_{mkd}$.
\textbf{5. Static MKD} removes $\mathcal{L}_{mkd}$, and performs mutual KD based on the performance of $R_1$, $R_2$, and $F$ evaluated at the very beginning of training stage two.
\textbf{6. Back KD} removes $\mathcal{L}_{mkd}$, and train $R_1$ and $R_2$ using knowledge distilled from $F$~\citep{Izacard2022FewshotLW}.

Results in Table~\ref{tab:ablation} show that the knowledge mining approach used in our model performs better than all other variants.
We can further observe that:
\textbf{1.} Retrieving in-context examples using other approaches (i.e., BM25 Retriever and EPR Retriever) degenerates the model performance by a large margin. This shows the effectiveness of the two-stage training of $R_1$ in our model.
\textbf{2.} Re-ranking hints generated by an LM help to improve the QA performance (see w/o Re-ranker).
\textbf{3.} Removing the adaptive mutual KD loss (i.e., w/o MKD) degenerates the QA performance. This proves the effectiveness of information exchange between the two branches of our model.
\textbf{4.} Variants of $\mathcal{L}_{mkd}$ lead to limited QA performance (see Static MKD and Back KD). This shows the importance of performance-aware for mutual knowledge distillation.

\subsection{Further Analysis}

\begin{table}[!t]
\centering
\small
\begin{tabular}{c|c|c c}
\toprule
Data & Methods & Tail@16 & $A_{\rm{unseen}}$\\

\midrule
 \multirow{2}{*}{\shortstack{w/o head\\ tasks}} & w/o $\mathcal{P}_m$ & 59.00 & 50.55 \\
& Ours & 59.54 (+0.54) & 51.05(+0.50)\\ 
\cmidrule{1-4}
 \multirow{2}{*}{\shortstack{w/ head \\tasks}} & w/o $\mathcal{P}_m$ & 59.56 & 52.09 \\
& Ours &  61.32 (\textbf{+1.76}) &  54.42(\textbf{+2.33})\\
\bottomrule
\end{tabular}
\caption{Effect of $\mathcal{P}_m$ in different data distributions.}
\label{tab:headsharing}
\end{table}

\paragraph{Effect of $\mathcal{P}_m$ in Different Data Distributions}
We also validate the effectiveness of meta prompt $\mathcal{P}_m$ for knowledge sharing in different data distributions.
Specifically, we construct a variant of the training set (and denote it as “w/o head”) by discarding samples from head tasks, which consist of samples from 16 tail tasks.
We also denote the original training set as “w/ head”.
The performance of our model on these two datasets is tested with and without $\mathcal{P}_m$.

Results in Table~\ref{tab:headsharing} show that our model benefits more from $\mathcal{P}_m$ with samples from head tasks.
This further validates our claim that meta prompt $\mathcal{P}_m$ helps to facilitate knowledge sharing between head, tail, and unseen tasks.

\begin{figure}[!t]
\scalebox{0.95}{
\centering
\includegraphics[width=1.0\linewidth]{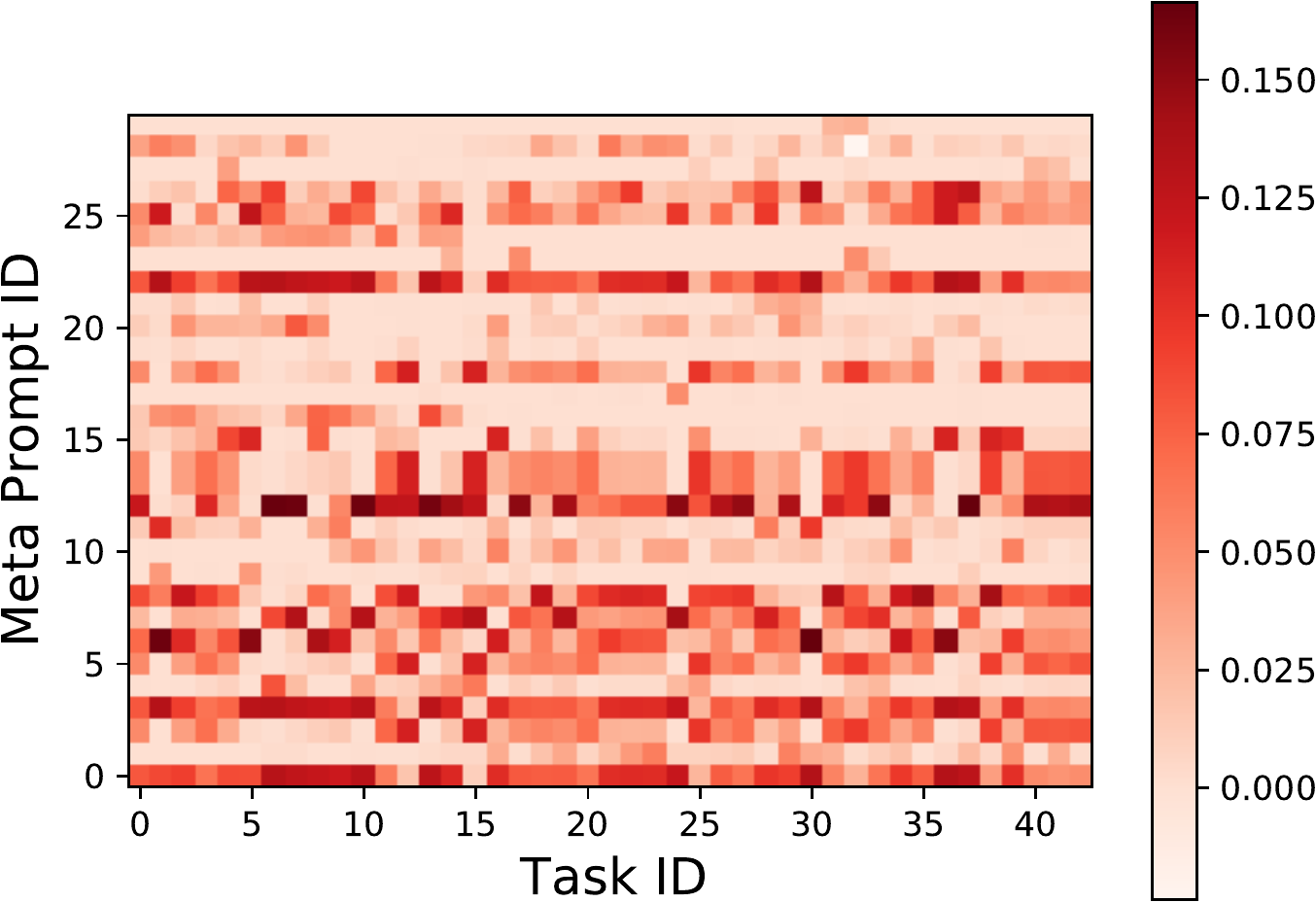}
}
\centering
\caption{Visualization of $\mathcal{P}_m$ selection mechanism.}
\label{fig:prompt-use}
\end{figure}

\paragraph{Analysis on $\mathcal{P}_m$ Selection Mechanism}
We plot the heat map of meta prompt $\mathcal{P}_m$ selection frequency for each task in Figure~\ref{fig:prompt-use}.
We can observe that:
\textbf{1.} Some hot meta prompts are shared by most tasks, which probably encode common knowledge for question answering.
\textbf{2.} Other meta prompts are shared by a few tasks, which might contain task-specific knowledge.

\begin{figure}[!t]
\scalebox{0.95}{
\centering
\includegraphics[width=1.0\linewidth]{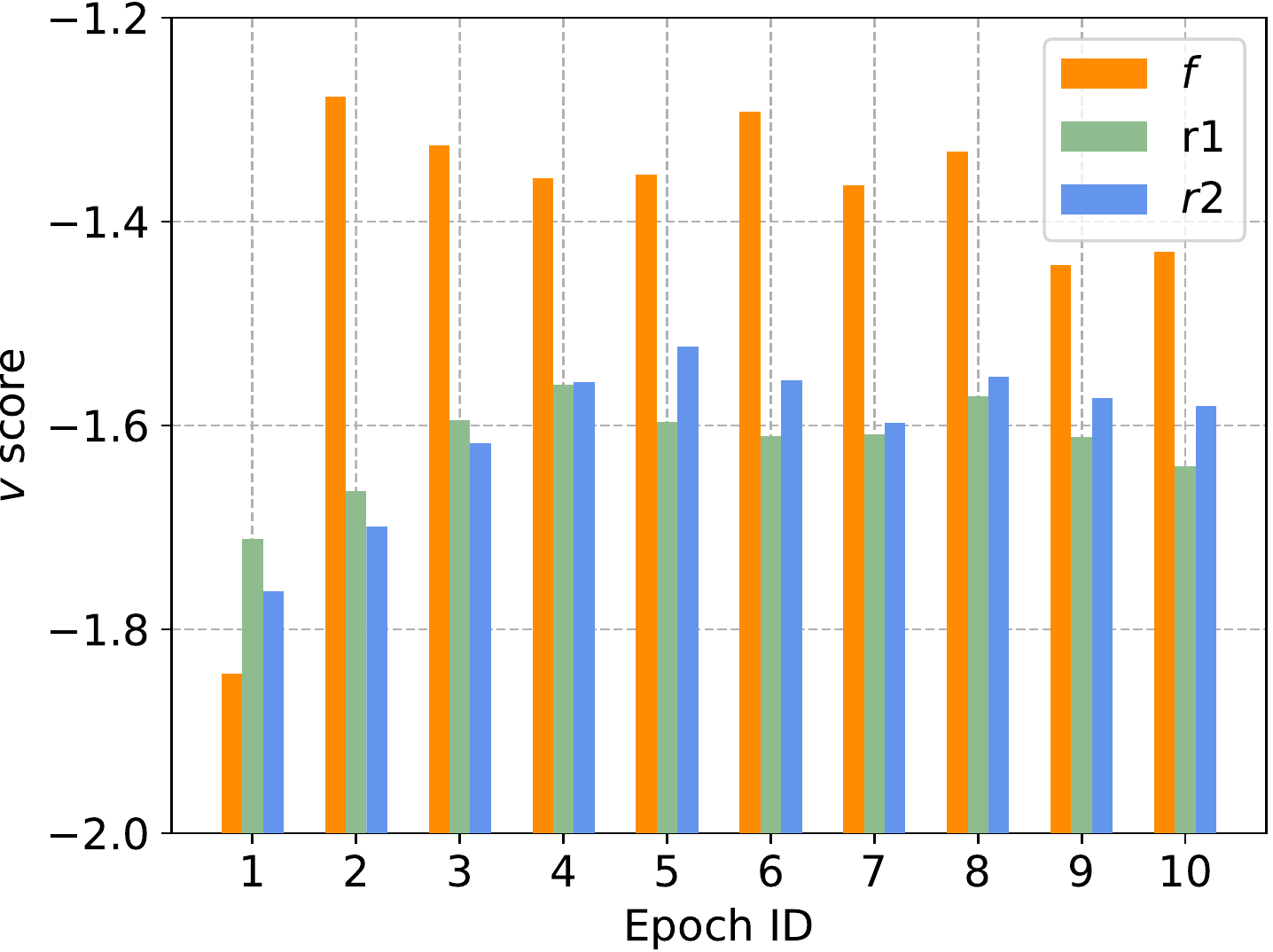}
}
\centering
\caption{Performance of retriever, re-ranker and the QA model in training stage two.}
\label{fig:module-score}
\end{figure}

\paragraph{Analysis on Adaptive Mutual KD} 
We visualize the performance of $R_1$, $R_2$, and QA model $F$ on the validation set $\mathcal{D}_{val}$ which are evaluated (Eq. \ref{evl}) at the beginning of each epoch during training stage two in Figure~\ref{fig:module-score}.
We can observe that:
\textbf{1.} Initially, $R_1$ and $R_2$ are allowed to generate knowledge for training $F$ because they are pre-trained in training stage one. After epoch one, $F$ performs better than $R_1$ and $R_2$, and starts to teach student model $R_1$ and $R_2$ as a teacher model.
\textbf{2.} During training, $R_2$ gradually outperforms $R_1$.
Overall, the relative performance of $R_1$, $R_2$, and QA model $F$ compared to each other is not stable during training. Thus, to avoid collective failures, being aware of individual performance is essential to perform mutual knowledge distillation.

\begin{figure}[t]
\centering
\subfigure[]{
\begin{minipage}[t]{0.5\linewidth}
\centering
\label{fig:alpha}
\includegraphics[width=1.0\linewidth]{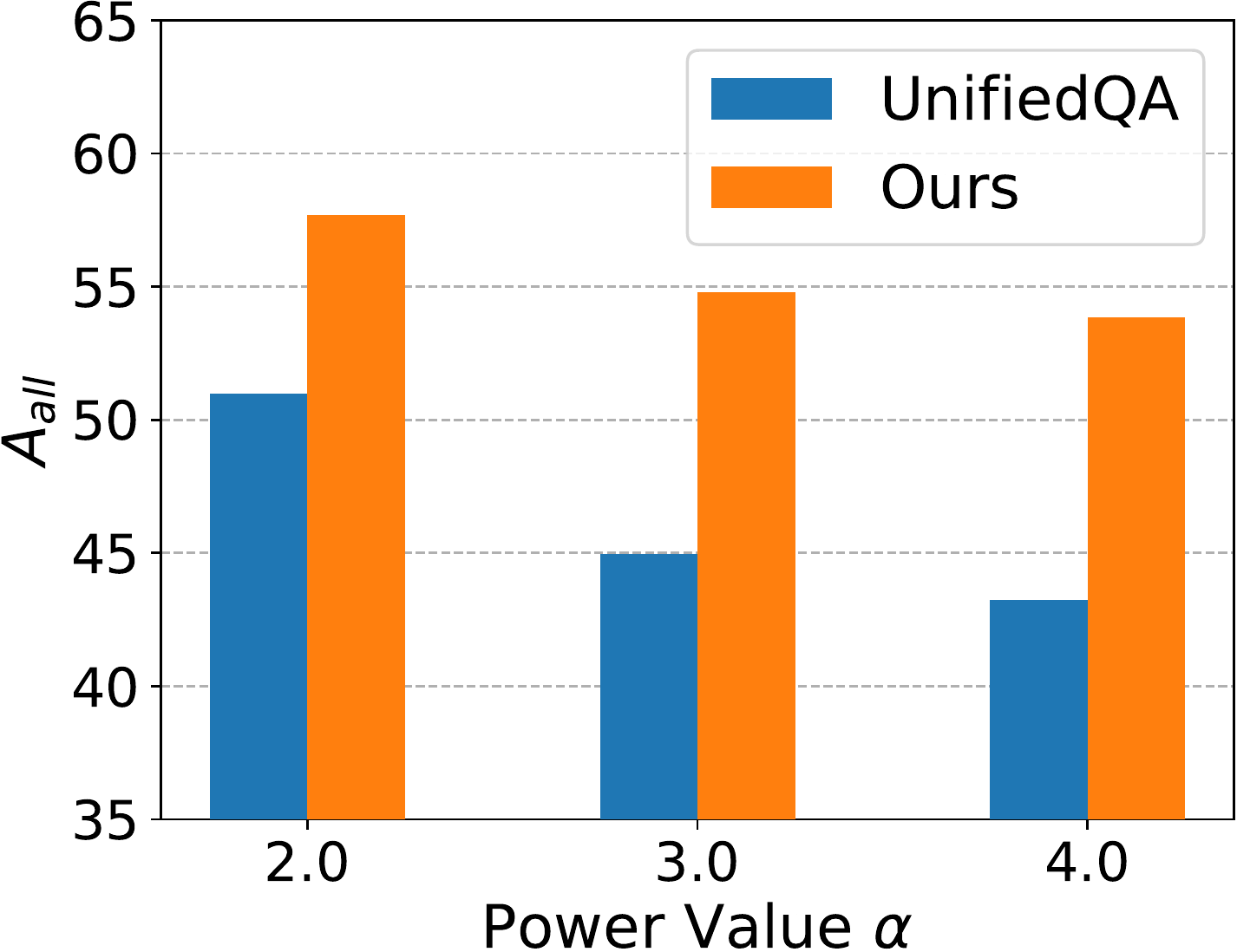}
\end{minipage}%
}%
\subfigure[]{
\begin{minipage}[t]{0.5\linewidth}
\centering
\label{fig:open}
\includegraphics[width=1.0\linewidth]{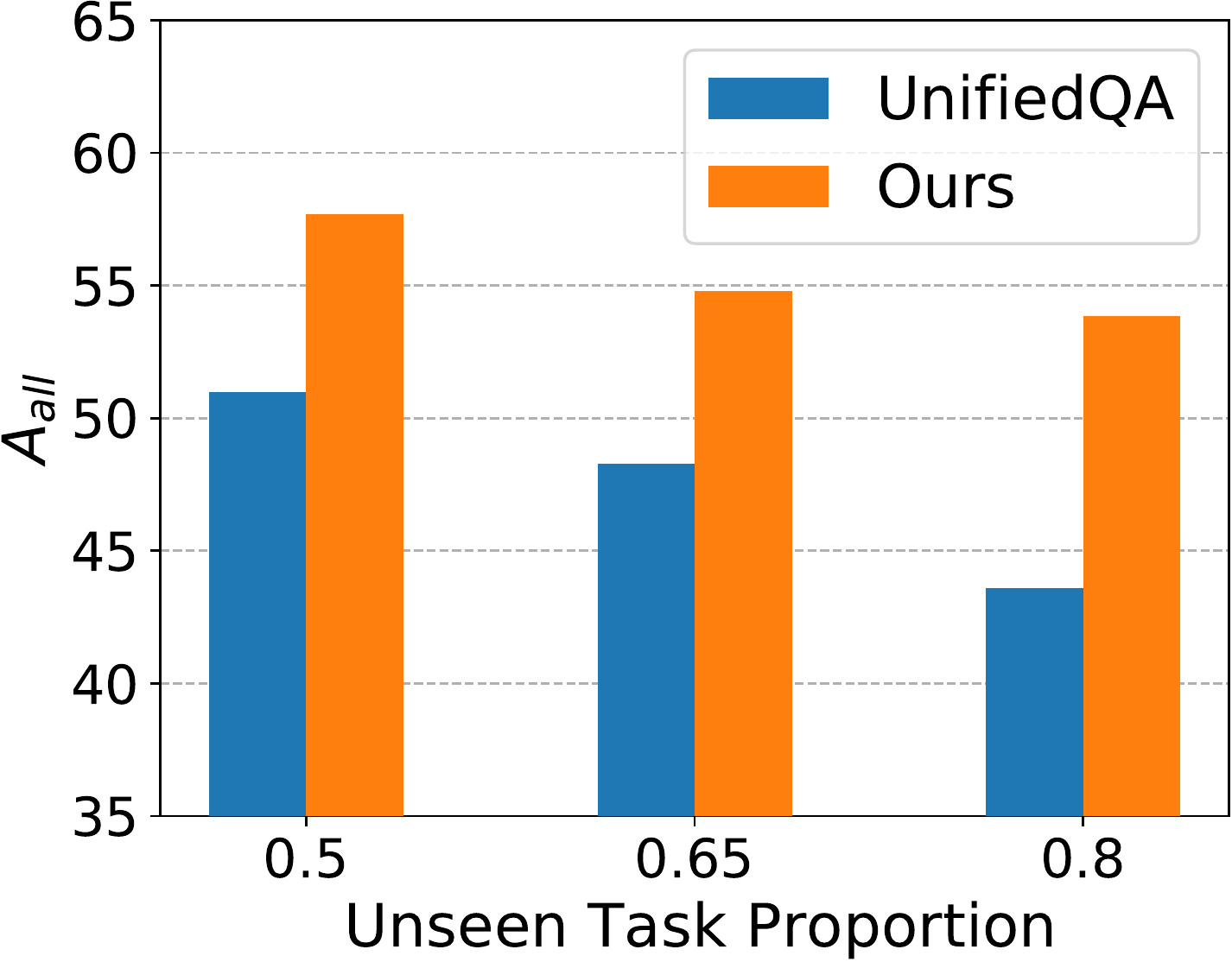}
\end{minipage}%
}%
\caption{The influence of (a) dataset longtail-ness and (b) proportion of unseen tasks over all 43 tasks.}
\centering
\label{fig:change-params}

\end{figure}

\paragraph{Influence of Dataset Longtail-ness}
The longtail-ness of the dataset (i.e., the degree of imbalance of task distribution in training) could have an impact on the model performance.
Figure~\ref{fig:alpha} shows that as the dataset becomes more imbalanced (i.e., $\alpha$ of Zipf distribution increases), our model only undergoes a moderate performance drop compared to UnifiedQA. Here, the performance is evaluated on a test set from all 43 tasks.

\noindent\textbf{Influence of Proportion of Unseen Tasks}
The performance change w.r.t. proportion of unseen tasks is shown in Figure~\ref{fig:open}.
Compared to UnifiedQA, the performance of our model changes steadily as the proportion of unseen tasks rises.
The knowledge sharing and knowledge mining components of our model enhance robustness to unseen tasks.



\section{Conclusion}
We introduce the open long-tailed QA (OLTQA) task that learns from natural long-tail distributed data and optimizes the performance over seen and unseen tasks.
We propose an OLTQA model to address the challenges of OLTQA.
An instance-level knowledge sharing mechanism is introduced, and a retrieve-then-rerank frame is employed to mine knowledge from a large pre-trained LM through a two-stage knowledge distillation training process.
We validate our model on a curated OLTQA benchmark.
Our publicly available data would enable future research that is directly transferable to real-world applications.

\section*{Limitations}
We identify the major limitation of this work is its input modality. Specifically, our model only considers textual inputs, ignoring question answering tasks in vision and audio. A multi-modal question answering model under realistic open long-tailed scenario is worth further exploration. Fortunately, through multi-modal pre-training models~\cite{xu-etal-2021-layoutlmv2,DBLP:journals/corr/abs-2103-06561} and question answering methods~\cite{Kim_2020_CVPR}, we can equip our model with multi-modal question answering ability. For future work, learning multi-modal question answering in an open (including out of distribution data~\citep{lang-etal-2022-estimating,lang2023out,lang2023survey}) long-tailed scenario still remains a challenge, and we will continue to work on it.
\section*{Ethics Statement}
This work does not raise any direct ethical issues.
In the proposed work, we seek to develop a method for long-tailed question answering in an open world, and we believe this work can benefit the field of question answering, with the potential to benefit other
fields involving open long-tailed problem. All experiments are conducted on open datasets.

\bibliography{anthology,custom}
\bibliographystyle{acl_natbib}

\appendix

\section{Datasets and Metrics}\label{append:datasets}
\noindent\textbf{Datasets.} We carry out experiments on the following datasets:
\begin{itemize}
    \item Extractive: SQuAD 1.1~\cite{rajpurkar-etal-2016-squad}, 
    SQuAD 2~\cite{rajpurkar-etal-2018-know}, 
    NewsQA~\cite{trischler-etal-2017-newsqa}
    Quoref~\cite{dasigi-etal-2019-quoref}, ROPES~\cite{lin-etal-2019-reasoning},
    AdversarialQA~\cite{bartolo-etal-2020-beat},
    ReCoRD~\cite{zhang2018record},
    \item Abstractive: DROP~\cite{dua-etal-2019-drop}
    NarrativeQA/NarQA~\cite{kocisky-etal-2018-narrativeqa}, the open-domain version of NaturalQuestions/NatQA~\cite{kwiatkowski-etal-2019-natural}, 
    QAConv~\cite{wu-etal-2022-qaconv}, 
    TweetQA~\cite{xiong-etal-2019-tweetqa}, 
    \item Multiple-choice: HeadQA~\cite{vilares-gomez-rodriguez-2019-head},
    RACE-C~\cite{pmlr-v101-liang19a},
    MCTest~\cite{richardson-etal-2013-mctest}, 
    RACE~\cite{lai-etal-2017-race}, OpenBookQA~\cite{mihaylov-etal-2018-suit}  ARC~\cite{clark2018think,clark2016combining},
    QASC~\cite{khot2020qasc},
    CommonsenseQA/CQA~\cite{talmor-etal-2019-commonsenseqa},
    Winogrande~\cite{sakaguchi2020winogrande},
    MMMLU~\cite{hendrycks2021measuring},
    ReClor~\cite{Yu2020ReClor:},
    Quail~\cite{rogers2020getting},
    OneStopQA~\cite{berzak-etal-2020-starc},
    MCScript~\cite{ostermann-etal-2018-mcscript},
    MCScript 2.0~\cite{ostermann-etal-2019-mcscript2},
    CosmosQA~\cite{huang-etal-2019-cosmos}, 
    ProcessBank~\cite{berant-etal-2014-modeling}, 
    DREAM~\cite{sun2019dream}, 
    PROST~\cite{aroca2021prost}, PhysicalIQA/PIQA~\cite{bisk2020piqa}, SocialIQA/SIQA~\cite{sap-etal-2019-social}
    \item Yes/no: BoolQ~\cite{clark-etal-2019-boolq},
    BoolQ-NP~\cite{khashabi-etal-2020-bang}
    the binary (yes/no) subset of MultiRC~\cite{khashabi-etal-2018-looking},  
    StrategyQA~\cite{geva-etal-2021-aristotle}, 
    PubmedQA~\cite{jin-etal-2019-pubmedqa}.
\end{itemize}
The statistics of these datasets are summarized in Table~\ref{tab:dataset}. Note that we follow the pre-process scheme released by~\citet{khashabi-etal-2020-unifiedqa} to tackle these datasets. As 22 tasks are unseen in the training phase, we only use the training and validation sets of the other 21 tasks to build our framework.

\noindent\textbf{Metrics.} 
The evaluation for each task follows~\citet{unifiedqav2}. Specifically, for Multiple-choice tasks, we use accuracy. For Extractive tasks, we use the F1 token overlap between the answer text and golden truth. For Abstractive tasks, we use ROUGE-L for NarrativeQA, BLEU for TweetQA, and F1 for the other tasks. For Yes/no questions, we also use the F1 token overlap.

\section{Overall Results}\label{ap:or}
We compare our OLTQA model with competitive baselines and ablation variants on each component. The full results of our model, baselines and ablation variants under 21 seen tasks are shown in Table~\ref{tab:in-domain-full}, while the results under 22 unseen tasks are shown in Table~\ref{tab:out-domain-full}. Bold numbers are superior results.

\begin{table*}[!t]
\centering
\small
\scalebox{0.8}{
\begin{tabular}{l|cccccccc}
\toprule
Methods & SQuAD 2 & NatQA & RACE & SQuAD 1.1 & DROP & NarQA & Winogrande & SIQA \\
\midrule
UnifiedQA & 77.80 & 40.25 & 56.97 & 85.32 & 32.50  & 44.69 & 54.93 & 50.15  \\
ProQA & 79.84 & 39.01 & 59.55 & 84.33 & 31.66  & 34.20 & 54.62 & \textbf{54.50}
\\
Muppet & 79.41  & 40.83 & 57.13  & 85.64 & 32.62 & 45.30  & 55.49 & 52.63 \\
Hyperformer++ & 79.52  & 40.24 & 58.24 & 87.13  & 32.17 & 51.88 & 54.93  & 52.46  \\
EPR & 44.14  & 39.50 & 38.82 & 87.12  & 29.22  & 46.02 & 51.70 & 45.96 \\
\midrule
Ours (w/o $\mathcal{P}_m$) & 77.72  & 42.10 & 58.13 & 85.98 & 35.53  & 56.89 & 54.85 & 49.64  \\
Ours (w/o $\mathcal{P}_k$) & 78.89  & 40.20 & 59.34 & 86.02  & 32.80  & 44.56 & 54.78 & 51.76  \\
Ours (w/o MKD) & 78.81 & 42.13 & 58.95 & 87.39 & 35.59  & 55.86 & 54.62 & 49.85  \\
Ours (BM25 Retriever)& 78.49 & 41.82 & 58.22 & 84.96 & 34.62  & 56.63 & 49.64 & 50.41
 \\
Ours (EPR Retriever) & 77.51  & 42.13 & 59.36 & 87.09 & 35.01   & 56.87 & 54.54 & 51.23  \\
Ours (w/o Re-ranker) & 77.94  & 41.50 & 57.64 & 86.73 & 34.54  & 56.04 & 55.56 & 50.67  \\
Ours (Static MKD)  & 78.73  & 42.67 & 59.55 & 87.72 & 35.81  & 57.34 & 55.33 & 51.48  \\
Ours (Back KD)  & 78.16  & 42.07 & 58.17 & 86.66 & 35.61  & 54.68 & 54.06 & 50.72  \\
Ours & \textbf{79.99}  & \textbf{42.68} & \textbf{59.65} &\textbf{87.88} & \textbf{36.42}  & \textbf{57.59} & \textbf{55.64} & 52.51 \\
\bottomrule
\toprule

Methods & Quoref & ROPES & CQA & BoolQ-NP & BoolQ & QASC & OBQA & PIQA \\
\midrule
UnifiedQA & 56.28 & 57.90 & 51.92 & 67.69 & 73.28 & 34.88 & 36.73  & 54.35 \\
ProQA & 35.75 & 30.10 & 51.52 & 69.67 & 72.51  & 31.10 & 43.40  & 56.31
\\
Muppet  & 57.66 & 55.42 & 53.79 & 68.84 & 74.27 & 32.62 & 39.47  & 55.47 \\
Hyperformer++ & 60.80 & 57.04 & 53.24  & 67.66 & 73.58  & 33.15 & 41.00 & 55.60 \\
EPR & 48.54 & 47.96 & 45.30 & 59.43 & 70.70  & 38.09 & 38.07  & 55.55 \\
\midrule
Ours (w/o $\mathcal{P}_m$) & 67.20 & 54.00 & 56.91 & 71.76 & 75.64 & 43.09 & 43.53  & 54.46 \\
Ours (w/o $\mathcal{P}_k$) & 56.32 & 57.96 & 52.50 & 70.64 & 74.62  & 36.83 & 39.53  & 55.98 \\
Ours (w/o MKD) & 69.00 & 52.66 & 55.61 & 71.77 & 76.18  & 46.00 & 43.80  & 55.22 \\
Ours (BM25 Retriever) & 68.09 & 54.10 & 52.66 & 71.07 & 72.84  & 42.76 & 39.00 & \textbf{56.43} \\
Ours (EPR Retriever) & 68.73 & 54.21 & 54.95 & 71.22 & 76.24  & 43.63 & 39.33  & 54.68 \\
Ours (w/o Re-ranker) & 65.38 & 53.28 & 52.83 & 72.18 & 73.17 & 39.52 & 39.67  & 53.70 \\
Ours (Static MKD)& 69.12 & 54.67 & 56.10 & 70.88 & 77.03 & 48.92 &  40.47  & 55.73 \\
Ours (Back KD)& 69.18 & 55.51 & 56.73 & 71.36 & 76.21 & \textbf{51.08} & 42.40   & 55.84 \\
Ours & \textbf{69.42} & \textbf{58.64} & \textbf{57.08} & \textbf{73.41} & \textbf{78.78}  & 50.65 & \textbf{44.27}  & 56.09
\\
\bottomrule
\toprule
Methods & NewsQA & ARC-easy & MCTest & ARC-hard & MultiRC  &  Head@5 & Tail@16 & $A_{\rm{seen}}$ \\
\midrule
UnifiedQA & 57.48 & 36.84 & 77.19 & 31.77 & 80.45 & 58.57 & 54.16 & 55.21  \\
ProQA & 49.93 & 44.21 & 80.00 & 38.13 & 77.56 & 58.88 & 51.47 & 53.23  \\
Muppet & 58.11 & 38.07 & 79.06 & 31.34 & 85.57 & 59.13 & 55.19 & 56.13  \\
Hyperformer++ & 59.45 & 40.18 & 76.88 & 31.10 & 86.86 & 59.46 & 55.99 & 56.81  \\
EPR & 18.26 & 51.81 & 55.00 & 39.80 & 56.41 & 47.76 & 48.04 & 47.97 \\
\midrule
Ours (w/o $\mathcal{P}_m$) & \textbf{59.70} & 56.49 & 83.02 & 39.46 & 85.58 &59.89 & 59.51 & 59.60  \\
Ours (w/o $\mathcal{P}_k$) & 58.87 & 39.82 & 76.25 & 33.11 & 85.90 & 59.45 & 55.59 & 56.51  \\
Ours (w/o MKD) & 58.88 & 57.37 & 82.19 & 39.46 & 84.94 &60.57 & 59.59 & 59.82  \\
Ours (BM25 Retriever) & 59.20 & 53.16 & 81.56 & 34.78 & 78.85 &59.62 &57.57 & 58.06  \\
Ours (EPR Retriever) & 58.99 & 56.49 & 81.98 & 36.12 & 83.65 & 60.22& 58.93 & 59.24  \\
Ours (w/o Re-ranker) & 59.49 & 51.58 & 80.94 & 37.15 & 87.18 &59.67 &58.02 & 58.41  \\
Ours (Static MKD) & 58.83 & 57.54 & 81.87 & 39.46 & 82.54 &60.90 & 59.83 & 60.09  \\
Ours (Back KD) & 58.87 & 57.89 & \textbf{85.63} & 40.22 & 83.18 & 60.13 & 60.24 & 60.21\\
Ours & 59.41 & \textbf{58.95} & 83.75 & \textbf{40.43} & \textbf{87.82} &\textbf{61.32} & \textbf{61.53} & \textbf{61.48}  \\
\bottomrule
\end{tabular}}
\caption{Comparison with competitive baselines and all ablations of our model in 21 seen tasks. Bold numbers are superior results.}
\label{tab:in-domain-full}
\end{table*}

\begin{table*}[!t]
\centering
\small
\scalebox{0.8}{
\begin{tabular}{l|cccccccc}
\toprule
\multirow{2}{*}{Methods} & AdversarialQA & AdversarialQA & AdversarialQA & \multirow{2}{*}{ReCoRD} & \multirow{2}{*}{RACE-C}  & \multirow{2}{*}{HeadQA} & \multirow{2}{*}{MMMLU} & \multirow{2}{*}{ReClor}  \\
& dBERT & dBiDAF & dRoberta  & & & & & \\
\midrule
UnifiedQA & 24.39 & 44.24 & 18.16 & 19.62 & 49.86 & 29.14 & 28.77 & 35.73 \\
ProQA & 24.13 & 41.67 & 14.21 & 13.42 & 54.91 & 29.84 & 25.96 & \textbf{37.60} \\
Muppet & 22.10 & 43.35 & 17.33 & 16.71 & 50.00 & 29.04 & 30.42 & 33.53 \\
Hyperformer++ & 20.09 & 45.30 & 16.99 & 17.74 & 52.11 & 28.62 & 25.26 & 35.47 \\
EPR & 37.00 & 53.76 & 27.74 & 8.98 & 35.39 & 32.21 & 28.77 & 25.07 \\
\midrule
Ours (w/o $\mathcal{P}_m$) & 34.51 & 51.42 & 25.16 & 13.76 & 53.51 & 34.55 & 33.68 & 33.73 \\
Ours (w/o $\mathcal{P}_k$) & 24.29 & 43.71 & 17.12 & \textbf{19.03} & 53.23 & 29.36 & 31.23 & 32.60 \\
Ours (w/o MKD) & 32.94 & 52.86 & 24.54 & 13.72 & 49.30 & \textbf{35.14} & 32.63 & 35.40 \\
Ours (BM25 Retriever) & 35.10 & 53.57 & 25.96 & 11.15 & 50.14 & 32.87 & 32.98 & 32.67 \\
Ours (EPR Retriever) & 37.26 & 54.58 & 26.80 & 14.11 & 53.65 & 34.00 & 32.72 & 34.73 \\
Ours (w/o Re-ranker) & 36.93 & 53.99 & 27.33 & 15.55 & 53.65 & 32.77 & 31.93 & 35.80 \\
Ours (Static MKD) & 32.47 & 53.13 & 24.89 & 13.80 & 54.21 & 35.07 & 34.39 & 32.93 \\
Ours (Back KD) & 31.66 & 53.91 & 24.91 & 15.64 & 53.14 & 35.00 & 32.63 & 34.89 \\
Ours & \textbf{39.51} & \textbf{55.12} & \textbf{28.05} & 17.97 & \textbf{56.88} & 34.48 & \textbf{36.14} & 36.67 \\ 
\bottomrule
\toprule
\multirow{2}{*}{Methods}  & \multirow{2}{*}{Quail} & OneStopQA & OneStopQA & OneStopQA & \multirow{2}{*}{MCScript} & MCScript & \multirow{2}{*}{CosmosQA} & \multirow{2}{*}{DREAM}   \\
& & elementary & intermediate  & advanced & & 2.0 & &\\ 
\midrule
UnifiedQA & 53.31 & 53.09 & 55.25 & 54.01 & 67.97 & 77.38 & 37.42 & 59.56 \\
ProQA & 54.16 & 62.35 & 62.65 & 61.11 & 71.23 & 76.44 & 39.23 & 64.41 \\
Muppet & 52.86 & 54.33 & 56.17 & 54.79 & 70.91 & 76.97 & 35.75 & 58.61 \\
Hyperformer++ & 54.09 & 54.63 & 55.86 & 59.88 & 71.51 & 76.62 & 37.35 & 59.31 \\
EPR & 41.29 & 63.58 & 58.95 & 60.49 & 65.56 & 63.56 & 38.66 & 53.92 \\
\midrule
Ours (w/o $\mathcal{P}_m$) & 56.17 & 60.19 & 62.96 & 61.11 & 77.46 & 76.88 & 45.09 & 68.28 \\
Ours (w/o $\mathcal{P}_k$) & 52.94 & 56.67 & 57.72 & 56.70 & 70.80 & 77.57 & 39.87 & 60.29 \\
Ours (w/o MKD) & 55.43 & 54.32 & 57.41 & 54.32 & 75.69 & 78.22 & 45.46 & 67.35 \\
Ours (BM25 Retriever) & 55.06 & 58.64 & 58.02 & 58.95 & 78.03 & 79.65 & 45.36
& 68.71 \\
Ours (EPR Retriever) & 55.20 & 60.80 & 60.49 & 60.19 & 76.97 & 76.98 & 45.96
&69.17 \\
Ours (w/o Re-ranker) & 52.98 & 59.57 & 55.25 & 57.10 & 74.49 & 77.48 & 45.03
& 64.75 \\
Ours (Static MKD) & 55.29 & 61.73 & 60.49 & 59.26 & 74.63 & 77.97 & 43.92 & 68.82 \\
Ours (Back KD) & \textbf{57.98} & 61.16 & 59.88 & 60.60 & 77.18 & \textbf{79.85} & 45.78 & 69.40 \\
Ours & 56.96 & \textbf{65.12} & \textbf{65.74} & \textbf{64.31} & \textbf{79.16} & 78.27 & \textbf{46.16} & \textbf{69.51}\\
\bottomrule
\toprule
Methods & ProcessBank & PROST & StrategyQA & PubmedQA  &QAConv &TweetQA & $A_{\rm{unseen}}$  \\
\midrule
UnifiedQA & 70.75 & 31.73 & 40.50 & 50.53  & 61.43 & 64.52 & 46.70 \\
ProQA & 69.39 & 31.30 & 49.96 & 58.00 & 59.73 & 63.83 & 48.27 \\
Muppet & 73.47 & 28.99 & 43.62 & 56.73 & 61.82 & 66.02 & 46.98 \\
Hyperformer++ & 72.79 & 32.34 & 49.52 & 53.00 & 58.93 & 61.44 & 47.22\\
EPR & 70.07 & 30.33 & 42.08 & 59.67 & 60.72 & 66.65 & 46.57 \\
\midrule
Ours (w/o $\mathcal{P}_m$) & 77.55 & 31.82 & 49.38 & 62.07 & 62.36 & 74.27 & 52.09
 \\
Ours (w/o $\mathcal{P}_k$) & 75.51 & 32.80 & 49.39 & 56.27 & 60.99 & 66.02 & 48.37
 \\
Ours (w/o MKD) & 74.83 & 31.66 & \textbf{51.44} & 61.60 & 62.18 & 73.33 & 50.90
\\
Ours (BM25 Retriever) & 75.28 & 31.43 & 51.35 & 58.93 & 61.39 & 76.44 & 51.44 \\
Ours (EPR Retriever) & 75.06 & 32.60 & 49.24 & 60.53 & 61.80 & 74.14 & 52.14 \\
Ours (w/o Re-ranker) & 73.02 & 29.80 & 51.31 & 61.60 & 62.26 & 69.53 & 51.01 \\
Ours (Static MKD) & 74.15 & 32.09 & 49.18 & 63.87 & \textbf{63.46} & 75.60 & 51.88 \\
Ours (Back KD) & 74.68 & 30.81 & 51.40 & 62.73 & 63.39 & 75.18 & 52.35 \\
Ours & \textbf{78.91} & \textbf{33.68} & 50.70 & \textbf{64.40} & 62.28 & \textbf{77.17} & \textbf{54.42} \\
\bottomrule
\end{tabular}}
\caption{Comparison with competitive baselines and all ablations of our model in 22 unseen tasks. Bold numbers are superior results.}
\label{tab:out-domain-full}
\end{table*}

\section{Case Study}\label{apc:case}
We provide examples from tail and unseen tasks, where our model is correct and the variant without knowledge mining (i.e., w/o $\mathcal{P}_k$) is incorrect, together with top hints selected by the retrieve-then-rerank framework.
Table~\ref{tab:cases-full} demonstrates that hints yielded by our model are related to the ground truth which effectively corrects the predicted answer.

\begin{table*}[!t]
\centering
\small
\scalebox{0.9}{
\begin{tabular}{c|c|c}
\toprule
Task    & Ours (w/o $\mathcal{P}_k$)  & Ours \\
\midrule
\multirow{7}{*}{NarQA} & \multicolumn{2}{l}{\textbf{Input}:The play begins with three...WHAT SENTENCE DID CYNTHIA GIVE TO THE SYMBOLIC VICES?}\\
& \multicolumn{2}{l}{\textbf{Ground Truth}:Make reperations and purify themselves.}\\
\cmidrule{2-3}
& \multirowcell{5}[0pt][l]{\shortstack{\textbf{Output}:To make reparation and purify themselves by \\ bathing in the spring.}} & \makecell[l]{\textbf{Hints}: To make reparation and to purify yourselves; } \\
&  & \makecell[l]{ Make reparation and to purify themselves by bathing in}  \\
& & \makecell[l]{the spring at Mount Helicon.; Make reparation and purify} \\
&  & \makecell[l]{yourselves.; Make reparation and purge yourselves}  \\
&  & \makecell[l]{\textbf{Output}:Make reparation and purify themselves} \\
\midrule
\multirow{5}{*}{ARC-hard} & \multicolumn{2}{l}{\textbf{Input}:A daphnia population... To which factor is the daphnia population most likely responding? (A) the pH of...}\\
& \multicolumn{2}{l}{\textbf{Ground Truth}:the temperature of the water}\\
\cmidrule{2-3}
& \multirowcell{3}[0pt][l]{\shortstack{\textbf{Output}:the pressure of the water }} & \makecell[l]{\textbf{Hints}: light intensity; temperature;  the temperature;} \\
&  & \makecell[l]{ the temperature of the water. }  \\
&  & \makecell[l]{\textbf{Output}:the temperature of the water} \\
\midrule
\multirow{5}{*}{NewsQA} & \multicolumn{2}{l}{\textbf{Input}:RIO DE JANEIRO, Brazil (CNN) -- A Brazilian supreme court judge...When did the mother die?}\\
& \multicolumn{2}{l}{\textbf{Ground Truth}:September}\\
\cmidrule{2-3}
& \multirowcell{3}[0pt][l]{\shortstack{\textbf{Output}:June 2004}} & \makecell[l]{\textbf{Hints}:in September; September.; during childbirth; }  \\
& & \makecell[l]{to David Goldman.} \\
&  & \makecell[l]{\textbf{Output}:September} \\
\midrule
\multirow{5}{*}{MultiRC} & \multicolumn{2}{l}{\textbf{Input}:German art collector...Was the Gurlitt art collection returned after confiscation?}\\
& \multicolumn{2}{l}{\textbf{Ground Truth}:yes}\\
\cmidrule{2-3}
& \multirowcell{3}[0pt][l]{\shortstack{\textbf{Output}:no}} & \makecell[l]{\textbf{Hints}: the surviving paintings were all returned; part}  \\
& & \makecell[l]{ of the collection was returned; part of it was; recently} \\
&  & \makecell[l]{\textbf{Output}:yes} \\
\midrule
\multirow{6}{*}{ReCoRD} & \multicolumn{2}{l}{\textbf{Input}:Lionel Messi is unattainable...Ariedo braida (pictured) says that it would be a mistake for \_ to change teams..}\\
& \multicolumn{2}{l}{\textbf{Ground Truth}:Lionel Messi}\\
\cmidrule{2-3}
& \multirowcell{4}[0pt][l]{\shortstack{\textbf{Output}:it would be a mistake for \_ to change teams}} & \makecell[l]{\textbf{Hints}: Barcelona; Lionel Messi is unattainable for}  \\
& & \makecell[l]{most football clubs; change teams; Messi is an icon of} \\
& & \makecell[l]{world football} \\
&  & \makecell[l]{\textbf{Output}:Lionel Messi} \\
\midrule
\multirow{6}{*}{TweetQA} & \multicolumn{2}{l}{\textbf{Input}:The way they run to each other... what does the tweeter imply?}\\
& \multicolumn{2}{l}{\textbf{Ground Truth}:they like each other}\\
\cmidrule{2-3}
& \multirowcell{4}[0pt][l]{\shortstack{\textbf{Output}:No Answer>}} & \makecell[l]{\textbf{Hints}: I had great time with my kids; they really like }  \\
& & \makecell[l]{each other; They want to know each other.; they are} \\
& & \makecell[l]{attracted to each other.} \\
&  & \makecell[l]{\textbf{Output}:they are attracted to each other.} \\
\midrule
\multirow{5}{*}{StrategyQA} & \multicolumn{2}{l}{\textbf{Input}:(Gulf of Finland) The bottom of...Would the Titanic be well preserved at the bottom of the Gulf of Finland?}\\
& \multicolumn{2}{l}{\textbf{Ground Truth}:yes}\\
\cmidrule{2-3}
& \multirowcell{3}[0pt][l]{\shortstack{\textbf{Output}:ships are relatively well preserved}} & \makecell[l]{\textbf{Hints}: yes; yes, it would be well preserved; Yes,}  \\
& & \makecell[l]{ it would.; well preserved} \\
&  & \makecell[l]{\textbf{Output}:yes} \\
\midrule
\multirow{5}{*}{RACE\_C} & \multicolumn{2}{l}{\textbf{Input}:Many post-80s...Many post-80s couples can't go to the movies, shop or attend parties because \_.? (A) they ...}\\
& \multicolumn{2}{l}{\textbf{Ground Truth}:they have to look after their kids}\\
\cmidrule{2-3}
& \multirowcell{3}[0pt][l]{\shortstack{\textbf{Output}:they have to look after their parents}} & \makecell[l]{\textbf{Hints}: their kids are born;  their kids were born;}  \\
& & \makecell[l]{ kids were born; they have to look after their kids} \\
&  & \makecell[l]{\textbf{Output}:they have to look after their kids} \\
\midrule
\bottomrule
\end{tabular}}
\caption{Case study from tail and unseen tasks where our model is correct and the variant without knowledge mining (i.e., w/o $\mathcal{P}_k$) is incorrect along with the top 4 hints selected by the retrieve-then-rerank framework.}
\label{tab:cases-full}
\end{table*}

\section{More Implementation Details}\label{app:details}
We use T5-base~\cite{raffel2020exploring} to initialize our encoder-decoder QA model (12 layers, 768-dimensional hidden size, and 12 attention heads). In knowledge sharing, we maintain totally $s=30$ meta prompts, and set the length of each meta prompt to $10$. We adopt a fixed T5-base encoder with an average pooling layer to generate the query vector. For each instance $\langle \bm{c},\bm{q},\bm{a}\rangle$, we select $\widetilde{s}=5$ meta prompts to construct $\mathcal{P}_m$. For meta prompt key training, we set $\eta = 0.15$ and $\gamma = 0.3$ in Eq.~\ref{eq:meta}. 

In knowledge mining, we adopt GLM-10B~\cite{du-etal-2022-glm} with 10B parameters as a large pre-trained LM. For retrieve-then-rerank example selection, $R_1$ first retrieves $l=64$ examples from all training examples, and $R_2$ selects $\widetilde{l}=4$ examples among retrieval results. The retriever $R_1$ is implemented with two separate dense encoders $E_X(\cdot)$ and $E_D(\cdot)$ to map $\langle\bm{c},\bm{q}\rangle$ and $\bm{e}_i$ into vectors. The score for $\bm{e}_i$ is then computed as $E_X([\bm{c};\bm{q}])^T\cdot E_D(\bm{e}_i)$, which is the dot product of two vectors. The re-ranker $R_2$ is a dense encoder $E_C$ combined with a linear layer $f_c$. Concretely, $E_C$ transforms the concatenation of example $\bm{e}_i$, hint $\bm{h}_i$ and input $\langle \bm{c},\bm{q} \rangle$ into a representation, which is fed into $f_c$ to get the score, denoted as $f_c(E_C([\bm{e}_i;h_i;\bm{c};\bm{q}]))$. $E_C$,$E_D$ and $E_X$ are all initialized with BERT base uncased~\cite{devlin-etal-2019-bert}. In two-stage training, we leverage BM25 to select $c=512$ example candidates. 

All experiments are performed on 8 A100 GPUs (80GB). The batch size is set to 32. We use the AdamW~\cite{loshchilov2017decoupled} optimizer with a learning rate of 1e-4 and batch size of 32. The dataset is trained for five epochs. All hyper-parameters are tuned according to the average score on the validation set. In our experiments, We perform 3 runs by setting the random seed to $\{42,43,44\}$ respectively. In this way, we report the average score of each method. Note that we only use the random seed $42$ for tuning hyper-parameters. Our model has 551.59M tunable parameters.

To obtain the ROUGE-L score, we use the NLTK package for sentence tokenization, and python rouge-score package for evaluation. To obtain the BLEU score, we use the NLTK package for evaluation.

\begin{table}[!t]
\centering
\small
\scalebox{0.8}{
\begin{tabular}{c|c|c|c|c}
\toprule
Format    & Dataset  & Train set size & Val set size & Test set size \\
\midrule
\multirow{9}{*}{Extractive} & SQuAD1.1 & 7978  & 886 & 10570 \\
          & SQuAD2 & 127319 & 3000 & 11873 \\
       & NewsQA    & 436  & 54 & 4341   \\
     & Quoref     & 1539 & 192  & 2768 \\
     & ROPES & 1242 & 155 & 1688\\
     & AdversarialQA(dBERT) & - & - & 1000\\
     & AdversarialQA(dBiDAF) & - & - & 1000 \\
     & AdversarialQA(dRoberta) & - & - & 1000 \\
     & ReCorD & - & - & 9999 \\
\midrule
\multirow{5}{*}{Abstractive}    & NarQA   & 3487 & 435 & 6922  \\
 & NQOpen & 31843 & 3980 & 10693\\
 & Drop & 5095 & 636 & 9536\\
 & QAConv & - & - & 3414 \\
 & TweetQA & - & - & 1086 \\
\midrule
\multirow{24}{*}{Multiple-choice}  & RACE      & 14205     & 1775 & 4887 \\
 & OBQA     & 566  & 70 & 500 \\
   & MCTest     & 335 & 41 & 320 \\
   & ARC-easy & 386 & 48 & 570 \\
   & ARC-hard & 309 & 38 & 299 \\
   & CQA & 1011 & 126 & 1221 \\
   & QASC & 638 & 79 & 926 \\
   & PIQA & 482 & 60 & 1838  \\
   & SIQA & 2031 & 253 & 1954 \\
   & Winogrande & 2573 & 321 & 1267 \\
   & RACE-C & - & -  & 712 \\
   & HeadQA & - & - & 1366 \\
   & MMMLU & - & - & 285 \\
   & ReClor & - & - & 500 \\
   & QuAIL & - &- & 2163 \\
   & OneStopQA elementary & - & - &  324 \\
   & OneStopQA intermediate & - & - &  324 \\
   & OneStopQA advanced & - & - &  324 \\
   & MCScript & - & - & 1411 \\
   & MCScript 2.0 & - & - & 2020 \\
   & CosmosQA & - & - & 2985 \\
   & ProcessBank & - & - & 147 \\
   & DREAM & - & - & 2040 \\
   & PROST & - & - & 18736 \\
\midrule
\multirow{5}{*}{Yes/no}  & BoolQ  & 748 & 93 & 3270 \\
&  MultiRC & 284 & 28 & 312 \\
& BoolQ-NP & 899 & 112 & 7596 \\
& StrategyQA & - & - & 2290 \\
& PubmedQA & - & - & 500 \\
\bottomrule
\end{tabular}}
\caption{Dataset Statistics.}
\label{tab:dataset}
\end{table}
\section{Results under Different Random Seeds}
We use random seed $42$ and $43$ to construct another two sets of head, tail, and unseen tasks, and compare our method with the baseline UnifiedQA. As shown in Table~\ref{tab:different}, our method is robust when using different tasks as head, tail or unseen tasks.
\begin{table}[!t]
\centering
\small
\begin{tabular}{c|l|c|c|c|c}
\toprule
     Seed & Method &Head@3& Tail@4 & $A_{\rm{seen}}$ & $A_{\rm{unseen}}$\\
\midrule
\multirow{2}{*}{42} 
& UnifiedQA & 49.68 & 56.54 & 47.74 & 40.19  \\
& Ours  & \textbf{53.10} & \textbf{66.29} & \textbf{56.03} &\textbf{49.76}  \\
\midrule
\multirow{2}{*}{43}
& UnifiedQA  & 56.71  & 50.05 & 50.65 & 42.67 \\
& Ours & \textbf{62.08} & \textbf{66.68} & \textbf{59.98} & \textbf{51.05} \\
\bottomrule
\end{tabular}
\caption{Results on different random seeds.}
\label{tab:different}
\end{table}
\end{document}